\documentclass[11pt]{article}

\pdfoutput=1
\usepackage[preprint]{latex/acl}

\usepackage{tikz}
\usepackage{tikz-3dplot}
\usetikzlibrary{3d,positioning, shapes.geometric, decorations.pathmorphing, backgrounds, fit}
\usetikzlibrary{shadows}

\usepackage{times}
\usepackage{latexsym}

\usepackage[utf8]{inputenc}
\usepackage[T5]{fontenc}

\usepackage{microtype}

\usepackage{inconsolata}

\usepackage{amssymb}
\usepackage{amsmath}
\usepackage{multirow}
\usepackage{adjustbox}
\usepackage{svg}
\usepackage{pgfplots}
\usepackage[disable]{todonotes}
\usepackage{booktabs}
\usepackage{float}
\usepackage{xspace}

\usepackage{graphicx}

\makeatletter
\DeclareRobustCommand\onedot{\futurelet\@let@token\@onedot}
\def\@onedot{\ifx\@let@token.\else.\null\fi\xspace}

\def\eg{\emph{e.g}\onedot} 
\def\ie{\emph{i.e}\onedot} 
 
 \def\vs{\emph{vs}\onedot}

\makeatother

\usepackage{amsmath,amsfonts,bm}

\def\eqref#1{equation~\ref{#1}}

\def\1{\bm{1}}

\def\ve{{\bm{e}}}

\def\vh{{\bm{h}}}

\def\vl{{\bm{l}}}
\def\vm{{\bm{m}}}

\def\vs{{\bm{s}}}

\def\vv{{\bm{v}}}

\def\mM{{\bm{M}}}

\DeclareMathAlphabet{\mathsfit}{\encodingdefault}{\sfdefault}{m}{sl}
\SetMathAlphabet{\mathsfit}{bold}{\encodingdefault}{\sfdefault}{bx}{n}
\newcommand{\tens}[1]{\bm{\mathsfit{#1}}}

\def\tC{{\tens{C}}}

\def\tL{{\tens{L}}}

\def\tR{{\tens{R}}}

\newcommand{\R}{\mathbb{R}}

\title{Scaling Graph-Based Dependency Parsing \\
  with Arc Vectorization and Attention-Based Refinement}

\author{Nicolas Floquet, Joseph Le Roux, Nadi Tomeh, Thierry Charnois \\
  Université Sorbonne Paris Nord, CNRS, \\Laboratoire d'Informatique de Paris Nord,\\
 F-93430 Villetaneuse, France\\
  \texttt{\{floquet,leroux,tomeh,charnois\}@lipn.fr} \\}

\begin{document}
\maketitle
\begin{abstract}

We propose a novel architecture for graph-based dependency parsing that explicitly constructs vectors, from which both arcs and labels are scored. Our method addresses key limitations of the standard two-pipeline approach by unifying arc scoring and labeling into a single network, reducing scalability issues caused by the information bottleneck and lack of parameter sharing. Additionally, our architecture overcomes limited arc interactions with  transformer layers to efficiently simulate higher-order dependencies. Experiments on PTB and UD show that our model outperforms state-of-the-art parsers in both accuracy and efficiency.

\end{abstract}

\section{Introduction}
\label{sec:introduction}

Recent graph-based dependency parsers have adopted a standard architecture \citep{kiperwasser2016simple,dozat-2017-deep-biaff} extended by \citet{zhang-etal-2020-efficient}. These models consist of two-pipelines: one pipeline scores arcs, and the other scores their labels.
Each pipeline uses independent models to generate specialized head and dependent representations from word embeddings, followed by a biaffine scoring model.

We investigate the \textit{scalability} of this widely-used architecture.
Our motivation stems from the observation that not all model architectures scale efficiently with increased parameters.
For example, transformer-based language models exhibit predictable scaling laws, where performance consistently improves with more parameters~\citep{kaplan2020scalinglawsneurallanguage}.
In contrast, other architectures, \eg{} CNNs, require careful scaling across multiple dimensions~\citep{efficientNet}.
Similar observations have been made in computer vision~\citep{dosovitskiy2021an}.
Our empirical results show that simply increasing the number of parameters in the standard parsing model does not improve performance.
We hypothesize that the core issue lies in the indirect representation of arcs.
The model encodes the entire space of possible arcs through word vectors and biaffine scoring, which limits its ability to handle increased complexity.
Furthermore, using two scoring networks restricts information flow between arc selection and labeling tasks.

We propose a novel architecture that explicitly constructs vector representations for each arc.
By unifying arc scoring and labeling tasks within a single network, our approach enables more effective parameter sharing and enhances scalability.
Additionally, to facilitate interaction between arc representations and efficiently simulate higher-order models, we employ transformer layers augmented with a differentiable filtering mechanism.
This design captures dependencies between arcs while maintaining computational and memory efficiency.



\section{Model}\label{sec:mod}

\begin{figure*}[htpb]
  \centering
\resizebox{1\textwidth}{!}{
\begin{tikzpicture}[thick]
\tikzstyle{block} = [rectangle, draw, rounded corners, minimum height=1em, minimum width=3cm, text centered, drop shadow]
\tikzstyle{biaffine} = [rectangle, draw, rounded corners, minimum height=1cm, minimum width=3cm, text centered, drop shadow]
\tikzstyle{triaffine} = [rectangle, draw, rounded corners, minimum height=1cm, minimum width=3cm, text centered, drop shadow]
\tikzstyle{transformer} = [rectangle, draw, rounded corners, minimum height=1cm, minimum width=3cm, text centered, drop shadow]
\tikzstyle{arrow} = [thick,->,>=stealth]
\tikzstyle{header} = [text centered, font=\large\bfseries]
\tikzstyle{innerblock} = [rectangle, draw, fill=purple!20, rounded corners, minimum height=1em, minimum width=3cm, text centered]
\tikzstyle{mlpblock} = [rectangle, draw, fill=purple!20, rounded corners, minimum height=1em, minimum width=1em, text centered, drop shadow]
\tikzstyle{cube} = [draw, fill=yellow!30, thick]

\node[header] (loc_title) at (-5.5, 0) {\textsc{Loc} and \textsc{CRF2o}};

\node (loc_arc_scores) [below left=0.3cm and 3.5cm of loc_title] {$s_{ij} \in \R$};
\node (loc_rel_scores) [right=4cm of loc_arc_scores] {$\vl_{ij} \in \R^\mathcal{L}$};
\node (c2o_sib_scores) [right=4cm of loc_rel_scores] {$s_{ijk} \in \R$};

\node[biaffine, fill=pink!20] (loc_arc_biaffine) [below=0.55cm of loc_arc_scores] {Biaffine $\mM \in \R^{a \times a}$};
\node[biaffine, fill=pink!20] (loc_rel_biaffine) [below=0.5cm of loc_rel_scores] {Biaffine $\tL \in \R^{c \times \mathcal{L} \times c}$};
\node[triaffine, fill=blue!20] (c2o_sib_triaffine) [below=0.5cm of c2o_sib_scores] {Triaffine $\tC \in \R^{d \times d \times d}$};

\node[mlpblock, fill=purple!20] (loc_mlp_arc_h) [below left=1.8cm and -1.25cm of loc_arc_biaffine] {Arc H};
\node[mlpblock, fill=purple!20] (loc_mlp_arc_m) [below right=1.8cm and -1.25cm of loc_arc_biaffine] {Arc M};

\node[mlpblock, fill=purple!20] (loc_mlp_rel_h) [below left=1.8cm and -1.25cm of loc_rel_biaffine] {Rel H};
\node[mlpblock, fill=purple!20] (loc_mlp_rel_m) [below right=1.8cm and -1.25cm of loc_rel_biaffine] {Rel M};

\node[mlpblock, fill=purple!20] (c2o_mlp_sib_h) [below left=1.8cm and -0.55cm of c2o_sib_triaffine] {Sib H};
\node[mlpblock, fill=purple!20] (c2o_mlp_sib_m) [below=1.8cm of c2o_sib_triaffine] {Sib M};
\node[mlpblock, fill=purple!20] (c2o_mlp_sib_s) [below right=1.8cm and -0.55cm of c2o_sib_triaffine] {Sib S};

\node[block, fill=blue!5, minimum width=12cm] (loc_bert) [below=4.5cm of loc_rel_biaffine] {Contextual Embedding};
\node[below=0.2cm of loc_bert, font=\large] (loc_inputcenter) {$\dots\quad w_j \quad\dots$};
\node[left=2.1cm of loc_inputcenter, font=\large] (loc_inputleft) {$\ldots\quad w_i$};
\node[right=2.1cm of loc_inputcenter, font=\large] (loc_inputright) {$w_k\quad \ldots$};

\begin{scope}[on background layer]
    \node[draw=gray, rounded corners, dashed, fit=(c2o_mlp_sib_h)(c2o_mlp_sib_m)(c2o_mlp_sib_s)(c2o_sib_triaffine)(c2o_sib_scores), inner xsep=0.4cm, inner ysep=0.2cm] (2nd_order_extension) {};
    \node[above=0cm of 2nd_order_extension.north] {\textit{2nd Order Extension}};
\end{scope}

\draw[arrow] (loc_arc_biaffine) -- (loc_arc_scores);

\draw[arrow] (loc_mlp_arc_h.north) to node[left, pos=0.5, xshift=0cm] {$\vh_i \in \R^a$} (loc_arc_biaffine);
\draw[arrow] (loc_mlp_arc_m.north) to node[right] {$\vm_j \in \R^a$} (loc_arc_biaffine);
\draw[arrow] (loc_rel_biaffine) -- (loc_rel_scores);
\draw[arrow] (loc_mlp_rel_h.north)  to node[left, pos=0.5, xshift=0cm] {$\vh_i \in \R^c$}  (loc_rel_biaffine);
\draw[arrow] (loc_mlp_rel_m.north) to node[right] {$\vm_j \in \R^c$}  (loc_rel_biaffine);
\draw[arrow] (c2o_sib_triaffine) -- (c2o_sib_scores);
\draw[arrow] (c2o_mlp_sib_h.north) to node[left, pos=0.5, xshift=0cm] {$\vh_i \in \R^d$} (c2o_sib_triaffine);
\draw[arrow] (c2o_mlp_sib_m.north) to node[] {$\vm_j \in \R^d$}  (c2o_sib_triaffine);
\draw[arrow] (c2o_mlp_sib_s.north) to node[right, pos=0.53] {$\vs_k \in \R^d\quad$}  (c2o_sib_triaffine);

\draw[arrow] (loc_bert.north) ++(-3.8, 0) -- ++(0, 0.5) node[above, name=loc_ei] {$\ve_i \in \R^b$};
\draw[arrow] (loc_bert.north) ++(0, 0) -- ++(0, 0.5) node[above, name=loc_ej] {$\ve_j \in \R^b$};
\draw[arrow] (loc_bert.north) ++(3.8, 0) -- ++(0, 0.5) node[above, name=c2o_ek] {$\ve_k \in \R^b$};
\draw[arrow] (loc_ei.north) -- (loc_mlp_arc_h.south);
\draw[arrow] (loc_ei.north) -- (loc_mlp_rel_h.south);
\draw[arrow] (loc_ei.north) -- (c2o_mlp_sib_h.south);
\draw[arrow] (loc_ej.north) -- (loc_mlp_arc_m.south);
\draw[arrow] (loc_ej.north) -- (loc_mlp_rel_m.south);
\draw[arrow] (loc_ej.north) -- (c2o_mlp_sib_m.south);
\draw[arrow] (c2o_ek.north) -- (c2o_mlp_sib_s.south);

\node[header] (arcloc_title) at (7, 0) {\textsc{ArcLoc} $P$\textsc{T}};

\node (arcloc_arc_scores) [below left=0.05cm and 0.2cm of arcloc_title] {$s_{ij} \in \R$};
\node (arcloc_rel_scores) [below right=0.05cm and 0.2cm of arcloc_title] {$\vl_{ij} \in \R^\mathcal{L}$};

\node[mlpblock, fill=purple!20] (arcloc_mlp_arc) [below=0.35cm of arcloc_arc_scores] {Arc Scoring MLP};
\node[mlpblock, fill=purple!20] (arcloc_mlp_rel) [below=0.3cm of arcloc_rel_scores] {Rel Scoring MLP};

\node[transformer, fill=yellow!30] (arcloc_transformer) at ($(arcloc_arc_scores)!0.5!(arcloc_rel_scores) - (0, 2.6)$) {$P \times$ Transformer Layers};

\begin{scope}[on background layer]
    \node[draw=gray, rounded corners, dashed, fit=(arcloc_transformer), inner sep=0.2cm] (option_transformer) {};
    \node[right=0.1cm of option_transformer, font=\bfseries] {Refinement};
\end{scope}

\node[biaffine, fill=pink!20] (arcloc_biaffine) [below=0.7cm of arcloc_transformer] {Biaffine $\tR \in \R^{d \times r \times d}$};

\node[mlpblock, fill=purple!20] (arcloc_arc_mlp_h) [below left=0.5cm and -1cm of arcloc_biaffine] {Head MLP};
\node[mlpblock, fill=purple!20] (arcloc_arc_mlp_m) [below right=0.5cm and -1cm of arcloc_biaffine] {Mod MLP};

\node[block, fill=blue!5, minimum width=5cm] (bert) at ($(arcloc_arc_mlp_h)!0.5!(arcloc_arc_mlp_m) - (0, 1.35)$) {Contextual Embedding};
\node[below=0.2cm of bert, font=\large] (inputcenter) {$\dots$};
\node[left=0.1cm of inputcenter, font=\large] (inputleft) {$\ldots\quad w_i$};
\node[right=0.1cm of inputcenter, font=\large] (inputright) {$w_j\quad \ldots$};

\draw[arrow] (bert) to node[left, pos=0.4] {$\ve_i \in \R^b\quad$}  (arcloc_arc_mlp_h.south) ;
\draw[arrow] (bert.40) to node[right, pos=0.4] {$\quad\ve_j \in \R^b$} (arcloc_arc_mlp_m.south);
\draw[arrow] (arcloc_mlp_arc) -- (arcloc_arc_scores);
\draw[arrow] (arcloc_transformer) to node[right, pos=0.6, xshift=0.75cm]{$\vv^P_{ij} \in \R^r$} (arcloc_mlp_arc);
\draw[arrow] (arcloc_biaffine) to node[right, pos=0.4] {$\vv^0_{ij} \in \R^r$} (arcloc_transformer);
\draw[arrow] (arcloc_arc_mlp_h.north) to node[left, pos=0.6] {$\vh_i \in \R^d\quad$} (arcloc_biaffine);
\draw[arrow] (arcloc_arc_mlp_m.north) to node[right, pos=0.6] {$\quad\vm_j \in \R^d$} (arcloc_biaffine);

\draw[arrow] (arcloc_mlp_rel) -- (arcloc_rel_scores);
\draw[arrow] (arcloc_transformer.40) -- (arcloc_mlp_rel);

\begin{scope}[on background layer]
    \draw[thick, gray] (2.9, 0) -- (2.9, -8);
\end{scope}

\end{tikzpicture}

}
  \caption{Illustration of both models. \textsc{Left:} standard model with 2 (resp. 3) pipelines for \textsc{Loc} (resp. \textsc{CRF2O}) with shared word embeddings. \textsc{Right:} our proposal with  a single pipeline and optionally $P$ transformers.}
  \label{fig:model}
\end{figure*}
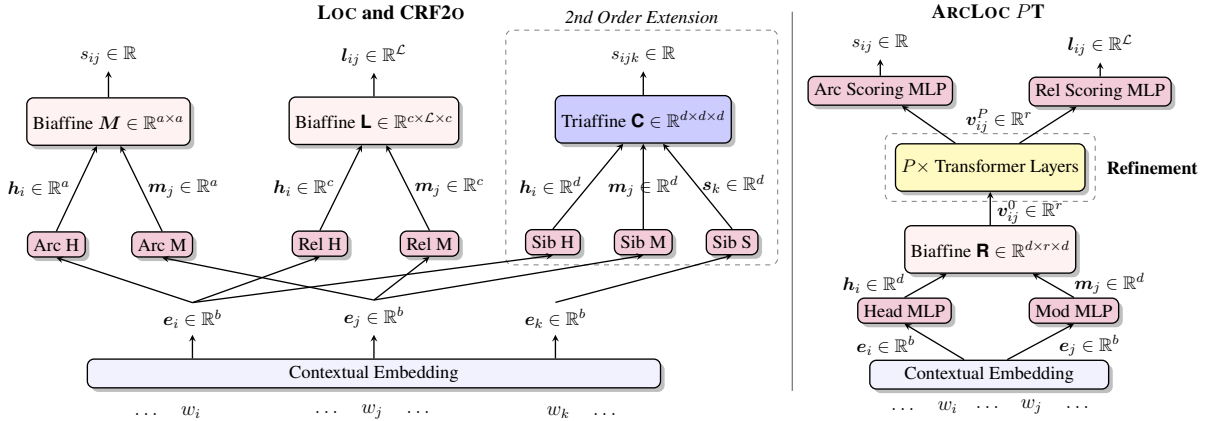

We review the standard biaffine parser (Figure \ref{fig:model}, left) and then highlight the key differences of our arc-centric approach (Figure \ref{fig:model}, right).
Prior to parsing, from an input sentence $x_{0}x_{1}\dots x_{n}$, where $x_{0}$ is the  dummy root and $\forall 1\leq i \leq n$, $x_{i}$ corresponds to the $i^{\text{th}}$ token of the sentence,  models start by computing contextual embeddings $\ve_{0}, \ve_{1}, \dots, \ve_{n}$.
This can be implemented in various ways, \eg{} with
averaged layers from pretrained dynamic word embeddings.
These contextual embeddings are further specialized for head and modifier roles using two feed-forward (FFN) transformations. This results in two sets of word representations, $\vh_{0}, \vh_{1}, \dots, \vh_{n}$ for heads and $\vm_{0}, \vm_{1}, \dots , \vm_{n}$ for modifiers.

\subsection{Standard Model}
We present the local and first-order models as introduced in~\cite{dozat-2017-deep-biaff} and refer readers to~\cite{zhang-etal-2020-efficient} for higher-order extensions.
The first-order scoring function decomposes the score of a parse tree as the sum of the scores of its arcs, if they form a valid tree (\ie{} rooted in $x_{0}$, connected and acyclic) and can be implemented as a CRF where arc variables are independently scored but connected to a global factor asserting well-formedness constraints.
This CRF can be trained efficiently and inference is performed with well-known algorithms.
Still, learning imposes to compute for each sentence \emph{its partition}, the sum of the (exponentiated) scores of all parse candidates.
While being tractable, this is an overhead compared to computing arc scores independently without tree-shape constraints.
Hence, several recent parsers, \eg{}~\cite{dozat-2017-deep-biaff} which called this model \emph{local}, simplify learning by casting it as a head-selection task for each word, \ie{} arc score predictors are trained without tree constraints.
In all cases, tree CRF or head selection, evaluation is performed by computing the optimal parse~\cite{eisner-1997-bilexical,tarjan-1977-findin}.

\textbf{Arc Scores} are computed by a biaffine function:\footnote{We ignore bias for the sake of notation.} for arc $x_{i}\to x_{j}$, \citet{dozat-2017-deep-biaff} set arc score to $s_{ij} =  \vh_{i}^{\top}\mM\vm_{j}$ with trainable $\mM$.
For embeddings of size $d$, $\mM$ has dimensions $d\times d$.

\textbf{Arc Labeling} is considered a distinct task: at training time arc labeling has its own loss and at prediction time most systems use a pipeline approach where first a tree is predicted, and second each predicted arc is labeled.\footnote{We remark that~\citet{zhang-etal-2021-strength} learn the two separately and merge them at prediction time.}
Labeling is also implemented with a biaffine:
for arc $x_{i}\to x_{j}$, the label logit vector is $\vl_{ij} = {\vh_{i}}^{\top}\tL\vm_{j}$, with trainable $\tL$.
For word vectors of size $d$ and for a system with $\mathcal{L}$ possible arc labels, $\tL$ has dimension $d\times \mathcal{L} \times d$.
While we noted them $\vh$ and $\vm$, these specialized word embeddings are given by FFNs  different from the ones used for arc scores.
This model relies on two biaffine functions, one for arc scores returning a scalar per arc, and one for labelings returning for each arc a vector of label scores.
Parameter sharing between them is limited to word embeddings $\ve$.

\subsection{Single Pipeline Model}

Our models differ architecturally  in two ways:
\emph{(i)} an intermediate vector representation is computed for each arc and \emph{(ii)} both arc and labeling scores are derived from this single arc representation.

For arc $x_{i}\to x_{j}$ we compute  vector representation $\vv_{ij}$.
Again, we use a biaffine function outputting a vector similarly to arc labeling in standard models:  $\vv_{ij} = \vh_{i}^{\top}\tR\vm_{j}$ for a trainable tensor $\tR$ with dimensions $d \times r \times d$, where $r$ is the size of the arc vector representation $\vv_{ij}$, and is a hyperparameter to be fixed, as is the word embedding size.
We recover arc score $s_{ij}$ and arc labeling $\vl_{ij}$ from $\vv_{ij}$
by FFNs: $s_{ij}=F_{s}(\vv_{ij})$ and $\vl_{ij}=F_{l}(\vv_{ij})$.
Note that there is only one biaffine function, and one specialization for head and modifiers.
Finally, remark that this change does not impact the learning objective: parsers are trained the same way.

\subsection{Refining with Attention}
\label{sec:transf-rerank}

Arc vectors obtained as above can read information from sentence tokens via contextual embeddings.
But we can go further and use Transformers~\cite{vaswani2017attention} to leverage attention in order to make arc representations aware of other arc candidates in the parse forest and adjust accordingly, effectively refining representations and realizing a sort of forest reranking.
We call \(\vv_{ij}^{0}\) the vector computed by the biaffine function over word embeddings described above.
Then we successively feed vectors of the form $\vv_{ij}^{p-1}$ to Transformer encoder layer $T^{p}$ in order to obtain $\vv_{ij}^{p}$ and eventually get the final representation $\vv_{ij}^{P}$.
This representation is the one used to compute scores with $F_{s}$ and $F_{l}$.
Remark again that this change in the vector representation is compatible with any previously used learning objectives.

The main issue with this model is the space complexity.
The softmax operation in Transformers requires multiplying all query/key pairs, the result being stored as a $t\times t$ matrix, where $t$ is the number of elements to consider.
In our context, the number of arc candidates is quadratic in the number of tokens in the sentence, so we conclude that memory complexity is $O(n^{4})$ where $n$ is the number of tokens.
To tackle this issue, we could take advantage of efficient architectures proposed recently \eg{}~Linear Transformers~\cite{qin-etal-2022-devil}.
Preliminary experiments showed training to be unstable so we resort to a filtering mechanism.

\paragraph{Filtered Attention} One way to tackle the softmax memory consumption is to filter input elements.
If the number of queries and keys fed to the transformer is linear, we recover a quadratic space complexity.
To this end we implement a simple filter $F_{f}$ to retrieve the best $k$ head candidates per word, reminiscent of some higher-order models prior to deep learning, \eg{}~\citet{koo-collins-2010-efficient} which used arc marginal probabilities to perform filtering.
We keep the $k$ highest-scoring  $F_{f}(\vv^{0}_{ij})$ for each position $j$, where $k$ typically equals $10$.
Kept vectors $\vv^{0}_{ij}$ are passed through the transformer as described above, while discarded ones are considered final.
This means that the transformer only sees arcs whose filter scores are among the highest-scoring ones,
the intuition being that transformers are only needed on cases where more context is required to further refine arc or label scores.
Implementation details can be found in Appendix~\ref{sec:filtering-arcs}.


\section{Experiments}\label{sec:exp}

\begin{table}[t]
  \centering
  \small
\begin{adjustbox}{width=\columnwidth}
\begin{tabular}{lrccrr}
                                                 & Speed & \multicolumn{2}{c}{Dev}         & \multicolumn{2}{c}{Test}        \\
                                                 &       & UAS            & LAS            & UAS            & LAS            \\ \midrule
\multicolumn{2}{l}{\citet{wang-tu-2020-second}$\star$}                      & -              & -              & 96.94          & 95.37          \\
\multicolumn{2}{l}{\citet{gan-etal-2022-dependency} Proj$\star$}        & -              & -              & 97.24          & 95.49          \\ \midrule
\multicolumn{2}{l}{\citet{yang-tu-2022-headed}$\star\star$}                & -              & -              & 97.4           & 95.8           \\
\multicolumn{2}{l}{\citet{amini-etal-2023-hexatagging} $\star\star$}       & -              & -              & 97.4           & 95.8           \\ \midrule
\multicolumn{6}{c}{\small \textit{4 million parameters}}                                                                              \\
\textsc{Loc}                                     & 353   & 96.85          & 95.16          & 97.36          & \textbf{95.90}          \\
\textsc{CRF2o}                                   & 144   & \textbf{96.87}          & \textbf{95.18}          & 97.33          & 95.89          \\
\textsc{ArcLoc} 0T                 & \textbf{356}   & 96.85          & 95.16          & \textbf{97.37}          & 95.86          \\
\textsc{ArcLoc} 1T                  & 337   & 96.84          & 95.13          & 97.36          & 95.81          \\
\textsc{ArcLoc} 2T                 & 329   & 96.81          & 95.12          & 97.35          & 95.82          \\ \midrule
\multicolumn{6}{c}{\small \textit{50 million parameters}}                                                                             \\
\textsc{Loc}                                     & \textbf{333}   & 96.83          & 95.16          & 97.36          & 95.91          \\
\textsc{CRF2o}                                   & 140   & 96.89          & 95.19          & 97.31          & 95.88          \\
\textsc{ArcLoc} 0T                 & \textbf{333}   & \textbf{96.91}          & \textbf{95.26}          & \textbf{97.37}          & 95.90          \\
\textsc{ArcLoc} 1T                  & 316   & 96.90          & 95.22          & 97.36          & 95.87          \\
\textsc{ArcLoc} 2T                 & 308   & 96.87          & 95.20          & \textbf{97.37}          & \textbf{95.91}          \\ \midrule
\multicolumn{6}{c}{\small \textit{100 million parameters}}                                                                            \\
\textsc{Loc}                                     & 301   & 96.79          & 95.12          & 97.35          & 95.87          \\
\textsc{CRF2o}                                   & 135   & 96.88          & 95.18          & 97.34          & 95.88          \\
\textsc{ArcLoc} 0T                 & \textbf{319}   & \textbf{96.92} & \textbf{95.29} & \textbf{97.38} & \textbf{95.92} \\
\textsc{ArcLoc} 1T                  & 292   & 96.91          & 95.23          & 97.35          & 95.86          \\
\textsc{ArcLoc} 2T                 & 283   & 96.90          & 95.22          & 97.34          & 95.85          \\ \midrule
\end{tabular}
\end{adjustbox}
\caption{Results on PTB test with RoBERTa, except for $\star\star$.
  $\star$: from \cite{gan-etal-2022-dependency}. $\star\star$: from \cite{amini-etal-2023-hexatagging}, using XLNet and no POS tags.}
  \label{tab:ptb}
\end{table}


\paragraph{Data}
We conduct experiments on the English Penn Treebank (PTB) with Stanford dependencies~\cite{de-marneffe-manning-2008-stanford}, as well as Universal Dependencies 2.2 Treebanks (UD; \citealt{11234/1-2837}), from which we select 12 languages, optionally pseudo-projectivized following~\cite{nivre-nilsson-2005-pseudo} for projective parsers.
We use the standard split on all datasets.
Contextual word embeddings are obtained from RoBERTa\textsubscript{large}~\cite{DBLP:journals/corr/abs-1907-11692} for the PTB and XLM-RoBERTa\textsubscript{large}~\cite{conneau-etal-2020-unsupervised} for UD.\@

\begin{table*}[t]
\begin{center}
\begin{adjustbox}{width=\textwidth}
\small
\setlength\tabcolsep{4pt}
\begin{tabular}{lrrlllllllllllll}
\multicolumn{2}{l}{Model} \#Param ($10^{6}$)
   &
  \multicolumn{1}{c}{Speed} &
  \multicolumn{1}{c}{bg} &
  \multicolumn{1}{c}{ca} &
  \multicolumn{1}{c}{cs} &
  \multicolumn{1}{c}{de} &
  \multicolumn{1}{c}{en} &
  \multicolumn{1}{c}{es} &
  \multicolumn{1}{c}{fr} &
  \multicolumn{1}{c}{it} &
  \multicolumn{1}{c}{nl} &
  \multicolumn{1}{c}{no} &
  \multicolumn{1}{c}{ro} &
  \multicolumn{1}{c}{ru} &
  \multicolumn{1}{c}{Avg} \\

  \midrule
\multicolumn{3}{l}{\cite{gan-etal-2022-dependency} Proj}         & 93.61          & 94.04          & 93.10          & 84.97          & 91.92          & 92.32          & 91.69          & 94.86          & 92.51          & 94.07          & 88.76          & 94.66          & 92.21 \\
\multicolumn{3}{l}{\cite{gan-etal-2022-dependency} NProj}     & 93.76          & 94.38          & 93.72          & 85.23          & 91.95          & 92.62          & 91.76          & 94.79          & 92.97          & 94.50          & 88.67          & 95.00          & 92.45 \\ \midrule

\textsc{VI}                              & 4   & 328 & 94.31          & 94.33          & 94.18          & 84.08          & 91.65          & 93.72          & 91.48          & 94.63          & 93.50          & 95.10          & 90.24          & 95.82           & 92.75          \\ \midrule

\textsc{Loc}                             & 4   & 497 & 94.54          & 94.60          & 94.15          & 85.54          & 92.36          & 93.96          & 91.70          & 95.18          & 94.14          & 95.34          & 90.27          & 95.79           & 93.13          \\
\textsc{Loc}                             & 50  & 463 & 94.41          & 94.53          & 94.15          & 85.28          & 92.19          & 93.88          & 91.72          & 95.11          & 94.06          & 95.19          & 90.16          & 95.80           & 93.04          \\
\textsc{Loc}                             & 100 & 426 & 94.37          & 94.49          & 94.11          & 85.25          & 92.21          & 93.81          & 91.75          & 95.09          & 93.96          & 95.18          & 90.21          & 95.80           & 93.02          \\ \midrule

\textsc{CRF2o}                           & 4   & 161 & 94.54          & 94.32          & 93.62          & 85.34          & 92.30          & 93.71          & 91.80          & 95.24          & 93.67          & 95.33          & 90.10          & 95.40           & 92.95          \\
\textsc{CRF2o}                           & 50  & 158 & 94.28          & 94.29          & 92.84          & 85.24          & 92.30          & 93.73          & 91.78          & 95.23          & 93.48          & 95.21          & 90.08          & 95.42           & 92.82          \\
\textsc{CRF2o}                           & 100 & 155 & 94.28          & 94.27          & 93.57          & 85.19          & 92.17          & 93.70          & \textbf{91.87} & 95.26          & 93.41          & 95.16          & 90.18          & 95.39           & 92.87          \\ \midrule

\textsc{ArcLoc 0T}          & 4   & 484 & 94.09          & 94.22          & 94.14          & 84.97          & 92.10          & 93.56          & 91.40          & 94.87          & 93.71          & 94.98          & 90.01          & 95.75           & 92.82          \\
\textsc{ArcLoc 0T}          & 50  & 459 & 94.33          & 94.50          & 94.28          & 85.35          & 92.35          & 93.94          & 91.78          & 95.06          & 94.03          & 95.27          & 90.32          & 95.83           & 93.09          \\
\textsc{ArcLoc 0T}          & 100 & 420 & 94.46          & 94.61          & \textbf{94.30} & 85.50          & 92.38          & 93.94          & 91.83          & 95.20          & 94.17          & 95.37          & 90.28          & 95.88           & 93.16          \\ \midrule

\textsc{ArcLoc 1T}          & 4   & 451 & 94.24          & 94.41          & 94.15          & 85.24          & 92.20          & 93.71          & 91.56          & 94.99          & 93.95          & 95.42          & 90.18          & 95.74           & 92.98          \\
\textsc{ArcLoc 1T}          & 50  & 421 & 94.47          & 94.72          & \textbf{94.30} & 85.52          & 92.43          & 94.01          & 91.71          & 95.30          & \textbf{94.22} & 95.63          & 90.34          & \textbf{95.89}  & 93.21          \\
\textsc{ArcLoc 1T}          & 100 & 393 & \textbf{94.56} & 94.76          & 94.29          & 85.62          & 92.44          & \textbf{94.07} & 91.80          & 95.29          & 94.18          & \textbf{95.71} & \textbf{90.38} & \textbf{95.89}  & \textbf{93.25} \\ \midrule

\textsc{ArcLoc 2T}          & 4   & 449 & 94.24          & 94.41          & 94.13          & 85.22          & 92.19          & 93.73          & 91.52          & 95.09          & 93.88          & 95.45          & 90.05          & 95.75           & 92.97          \\
\textsc{ArcLoc 2T}          & 50  & 419 & 94.53          & 94.72          & \textbf{94.30} & 85.60          & 92.41          & 94.02          & 91.75          & \textbf{95.34} & \textbf{94.22} & 95.65          & 90.32          & \textbf{95.89}  & 93.23          \\
\textsc{ArcLoc 2T}          & 100 & 387 & 94.55          & \textbf{94.79} & \textbf{94.30} & \textbf{85.68} & \textbf{92.46} & \textbf{94.07} & 91.78          & 95.26          & 94.11          & 95.64          & 90.32          & \textbf{95.89}  & 93.24          \\

\end{tabular}
\end{adjustbox}
\end{center}
\caption{Test LAS for 12 languages in UD2.2. $P$T is the number of transformer layers.
}\label{tab:ud}

\end{table*}


\paragraph{Evaluation}
We report unlabeled and labeled attachment scores (UAS/LAS), with the latter to select best models on validation.
Results are averaged over 8 randomly initialized runs.
Following~\citet{zhang-etal-2020-efficient} and others, we  omit punctuations when evaluating on PTB but keep them on UD.\@
Finally, we use gold POS on UD but omit them for PTB.\@

\paragraph{Models}
\textsc{Loc} is the local model from~\cite{zhang-etal-2020-efficient} trained with arc cross-entropy while \textsc{CRF2o} is their second-order CRF.\@
\textsc{Vi} is the non-projective second-order CRF implementing mean-field variational inference~\cite{wang-tu-2020-second}.
\textsc{ArcLoc} is our model with arc vectors trained with arc cross-entropy.
All models\footnote{Models are based on \url{https://github.com/yzhangcs/parser} and will be publicly available upon publication.} are evaluated with the Eisner algorithm~\cite{eisner-1997-bilexical} extended to higher-order for \textsc{CRF2o} on PTB.
For UD, we use the MST algorithm~\cite{mcdonald-etal:2005:hltemnlp} for all parsers but \textsc{CRF2o} for which we report deprojectized results.
We tested 3 parameter regimes: small (4M), big (50M) and large (100M).
Hyperparameter details are given in Appendix~\ref{sec:hyperparameters}.
We include recently published results for comparison.

\paragraph{Main Results}
Our results on PTB (Table~\ref{tab:ptb}) show that our approach is slightly faster and improves LAS on the dev set over \textsc{Loc} and other state-of-the-art parsers.
Increasing the number of parameters is beneficial for our model, detrimental for \textsc{Loc}, and has no significant effect for \textsc{CRF2o}.
We also remark that on PTB, arc interactions through higher-order scoring or transformer layers have no beneficial impact.

For the 12 tested UD languages Table~\ref{tab:ud} reports results where we can see that on 
11 languages out of 12 a configuration of our parser achieves better performance than \textsc{Loc}, \textsc{Vi}\footnote{We only report 4M for \textsc{Vi} since we found training to be unstable otherwise, leading to performance collapse.} and \textsc{CRF2o}.
We notice that on UD the use of transformers allows for better results.
By increasing the number of parameters in \textsc{ArcLoc} we manage to achieve state-of-the-art performances at little cost in parsing speed.

Detailed results on dev sets are given in Appendix~\ref{sec:details} and an error analysis in Appendix~\ref{sec:error_analysis}.


\section{Related Work}\label{sec:rel}

Our model, assigning vectors to arcs, \ie{} the objects to be scored, draws inspiration from the auto-regressive neural approach to parsing~\cite{dyer2015bilstm}, as well as from span-based parsers such as~\cite{stern-etal-2017-minimal,zhou-zhao-2019-head} and arc-hybrid parsing in~\cite{le-roux-etal-2019-representation}.
Refining initial arc representations has also been explored~\citep{strubell2017dilatedCNN,mohammadshahi-henderson-2021-recursive}.
Our model with transformers bears a resemblance to earlier work on forest reranking for parsing~\cite{collins-koo-2005-discriminative,le-zuidema-2014-inside},
and to \cite{ji-etal-2019-graph} where the parse forest is exploited to recompute vectors for words, not arcs.

Attention is widely utilized in parsing~\cite{mrini-etal-2020-rethinking,tian-etal-2020-improving}, possibly with ad-hoc constraints on attention \citep{kitaev-klein-2018-constituency}.

Representing spans has been shown to be beneficial for NLP~\cite{li-etal-2021-span,yan-etal-2023-joint,yang-tu-2022-headed} as well as using transformers to enhance them~\cite{zaratiana-etal-2022-gnner}.
Our method uses standard softmax attention with a differentiable filter as opposed to rigid constrained masking~\cite{DBLP:journals/corr/abs-2112-00578} and other forms of attention~\cite{nodeformer,kim2017structured,cai2019graph,Hellendoorn2020Global}.
Our model is part of the literature on generalizing transformers to relational graph-structured data~\citep{battaglia2018relational,kim2022pure,ying2021do}.


\section{Conclusion}

We presented a change in the main graph-based dependency parsing architecture where  arcs have their own vector representation, from which scores are computed.
Our model improves parsing metrics and achieves state-of-the-art results on PTB and 11 UD corpora. We also demonstrated that transformer-based refinement simulates higher-order interactions and enhances parameter scalability.
Our model can be extended to other tasks, such as constituent parsing or relation extraction.

\clearpage
\section{Limitations}
\label{sec:limitations}

Our system with Transformers relies on the attention mechanism which is quadratic in space and time in the number of elements to consider.
Since the number of elements (arcs in our context) is itself quadratic in the number of word tokens, this means that naively the proposed transformer extension is of quadratic complexity. In practice we showed that adding a filtering mechanism is sufficient to revert complexity back to $O(n^{2})$, but we leave using efficient transformers, with linear attention mechanism, to future work.

Our model requires more parameters than previously proposed architecture to achieve the same level of performance.
This might be an issue for memory limited systems.

\section{Ethical Considerations}
\label{sec:ethic-cons}

We do not believe the work presented here further amplifies biases already present in the datasets.
Therefore, we foresee no ethical concerns in this work.

\section{Acknowledgments}
\label{sec:acknowledgments}
This work was granted access to the HPC resources of IDRIS under the allocation 2023-AD011013732R1 made by GENCI. This work was supported by the Labex EFL (Empirical Foundations of Linguistics, ANR-10-LABX-0083), operated by the French National Research Agency (ANR).

\bibliography{main}

\appendix

\section{Hyperparameters}
\label{sec:hyperparameters}
We mostly use the same hyperparameter settings as \citet{zhang-etal-2020-efficient} which are found in their released code.\footnote{\url{https://github.com/yzhangcs/parser}}
Specifically we adopt the approach they use when training models using BERT, using the average of the 4 last layers to compute our word embeddings, and also using a batch size of 5000, the dropout rate for all of our MLPs is 0.33, we train our model for 10 epochs and save the one with the best LAS score on the dev data.

\paragraph{\textsc{Loc}} We use arc MLP output sizes of 900, 3750, 5500 and label MLP output sizes of 150, 750, 1100 for the small (4$\times10^6$ parameters), big (50$\times10^6$ parameters) and large (100$\times10^6$ parameters) models respectively.

\paragraph{\textsc{ArcLoc}} In the small model, the dimension of the arc MLP is 155 without any attention layers, and 150 when using 1 or 2 layers, the arc sizes are 160 when using 0 or 1 layer of attention and 155 when using 2.
In the big model, the arc MLP dimension is 500 and the arc size is 192 no matter the number of attention layers we use and for the large model, we increase these sizes to 625 and 256 respectively.

\paragraph{Transformer} Our transformer uses a number of attention heads as close to one sixteenth of the arc size as we can get while following the rule that the arc size must be a multiple of the number of attention heads.
The transformer in \textsc{ArcLoc} benefits from its own hyperparameters, while the model warms up for one epoch, the transformer does so for three and has a base learning rate of 2.5e-3, which becomes 1.35e-4 when using SWA.

\paragraph{Miscellaneous} The learning rates are 8.3e-5 and 3.7e-5 for \textsc{Loc} and \textsc{ArcLoc} respectively before the stochastic weight averaging (SWA) and 5e-6 and 3.7e-6 also respectively from the fifth epoch onward when we use SWA.

\paragraph{Other Parsers} For \textsc{CRF2o}, we start from the parameters as \citet{zhang-etal-2020-efficient} with a few changes, the learning rates which are the same as \textsc{Loc}, and we have 3 different MLP sizes for the 3 model sizes, for the small model, the sizes are 560, 112 and 112 for the arc, rel, and sib MLPs respectively, for the big model, they are 1675, 335, 335, respectively and for the large model, 2150, 430, and 430, respectively.
For \textsc{Vi}, we start with the released code of the implementation by \citet{zhang-etal-2020-efficient}, and apply the exact same changes we applied to \textsc{CRF2o}.

\paragraph{Parameter Count}
We use RoBERTa's and XLM-RoBERTa's contextual embeddings of size 1024.
Single layer MLPs to obtain $\vh,\vm$ vectors of size $o$ (ignoring bias term) contain $1024o$ parameters.
Biaffine layers (without bias) of input size $i$ and output size $o$ have $i^{2}o$ parameters.

Accordingly, we use the following formula to determine the parameter count for \textsc{Loc} with 2 arc MLPs, 2 label MLPs, and 2 biaffine modules, one for the arcs and one for the labels:
\begin{align*}
  & 2\times 1024x + 2\times1024y + x{^2} + y{^2}\mathcal{L}\\
 =& 2048(x+y) + x{^2} + y{^2}\mathcal{L}
\end{align*}
where $x,y$ are the arc and label MLP output dimensions respectively and $\mathcal{L}$ is the number of labels in the dataset.

For \textsc{ArcLoc}, we use 2 single-layer MLPs for $\vh,\vm$ with output size $d$ and one biaffine layer of input size $d$ and output size $r$.

We also use 2 MLPs with a hidden layer to compute arc scores and labeling scores. These MLPs with input size $r$, hidden size $\frac{r}{2}$ for arcs and $2\mathcal{L}$ for labels, 
and output size either 1 for scores and $\mathcal{L}$ for labels respectively contain $r\times\frac{r}{2} + \frac{r}{2}$ and $r\times 2\mathcal{L} + 2\mathcal{L} \times \mathcal{L}$ parameters.

\begin{align*}
  &2\times 1024d + d{^2}r + r\frac{r}{2} + \frac{r}{2}  + 2\mathcal{L} \times (r + \mathcal{L})\\
 =&2048d + d{^2}r + \frac{r}{2}(1+r) + 2L(r+L)
\end{align*}

Additionally, each layer of Transformer adds (attention + MLP with hidden layer):
\begin{align*}
r^{2} + r\times (4r) + (4r)\times r =r^{2} + 8r^{2} = 9r^2
\end{align*}

\textsc{CRF2o} and \textsc{VI} require to add 3 single-layer MLPs with output size $z$ and a triaffine layer for sibling scores with output size 1, on top of the \textsc{Loc} parameters:

\begin{align*}
3072z + z^{3}
\end{align*}

\section{Stochastic Weight Averaging}

We implement stochastic weight averaging (SWA) introduced in \citet{61aa9e9cc965421e82d7b7042c61abc8} after 4 epochs, which we found lead to consistent improvements in all models (\textsc{Loc}, \textsc{ArcLoc}, \textsc{CRF2o}) after finetuning.

\section{Filtering Arcs}
\label{sec:filtering-arcs}

The filtering step keeps $k$ arcs per modifier.
It is inspired from the straight-through estimator~\cite{bengio2013estimating} and is implemented as follows.

For each token $m$ we compute the scores of all arcs $h \to m$, from their vector representations $v_{hm}$.
Then we add some Gumbel noise (at training time) and normalize scores via softmax: we obtain probabilities $p(h \to m)$ that we use to sort arcs from  most to least probable: $h_{1} \to m \dots h_{n} \to m$.

Finally the $k^{\text{th}}$ arc vector returned by the filter for modifier $m$ is computed as:

\begin{multline*}
  v_{k}(m) = \text{argsort}(v_{h_{1}m}\dots v_{h_{n}m})[k] - \\   \text{detach}(\mathbb{E}_{p(\cdot\to m)}[v_{hm}]) + \mathbb{E}_{p(\cdot\to m)}[v_{hm}]
  \end{multline*}

During the forward pass the two last terms cancel each other out and $v_{k}(m)$ is the vector of the $k^{\text{th}}$ most probable arc for $m$, $h_{k} \to m$.
During the backward pass, the first two terms have zero gradient, and the third one amounts to  a weighted average of the vectors of arcs $h_{1} \to m\dots h_{n}\to m$, with weights given by their probabilities.

Table~\ref{table:oracle_score} compares parsing UAS and the filter's oracle UAS (percentage of correct heads in the set returned by the filter). We keep 10 potential heads per word to get the highest oracle score with a reasonably small sequence of arcs.\footnote{Note that there is no discrepancy in the first or second column, we can have a UAS score higher than filter's oracle, as an arc can be filtered out and still end up in the parse, our filter only chooses arcs to be processed by the transformer.}
\begin{table}[H]
  \centering
  \small
\begin{tabular}{llllll}
  \multicolumn{1}{c}{\#Heads} & \multicolumn{1}{c}{1} & \multicolumn{1}{c}{2} & \multicolumn{1}{c}{3} & \multicolumn{1}{c}{5} & \multicolumn{1}{c}{10} \\
  \midrule
Oracle            & 37.65       & 75.88     & 92.48       & 99.10       & 99.88                       \\
Parser            & 48.79       & 78.06     & 89.69       & 94.74       & 96.88                       \\
\end{tabular}
\caption{PTB Dev UAS scores for \textsc{ArcLoc} 1T and its filter's Oracle with different filter sizes (number of kept heads per word).}
\label{table:oracle_score}
\end{table}


\section{UD Development Results}
\label{sec:details}

\begin{table*}[t]
\begin{adjustbox}{width=1\textwidth}
\centering
\begin{tabular}{lllllllllllllll}
\midrule
                               & \# Param ($10^{6}$) & bg  & ca  & cs  & de  & en  & es  & fr  & it  & nl  & no  & ro  & ru  & Avg            \\ \midrule
projective\%                   &   & 99.8    & 99.6    & 99.2    & 97.7    & 99.6    & 99.6    & 99.7    & 99.8    & 99.4    & 99.3    & 99.4    & 99.2    & 99.4           \\ \midrule

\textsc{Vi}                      & 4   & 92.93 & 94.09 & 94.51 & 88.44 & 92.43 & 93.91 & 92.86 & 94.04 & 94.78 & 95.56 & 90.19 & 95.27 & 93.25  \\ \midrule

\textsc{Loc}                     & 4   & 93.10 & 94.35 & 94.52 & 89.61 & 93.04 & 94.17 & 93.04 & 94.59 & 95.18 & 95.83 & 90.07 & 95.31 & 93.57  \\
\textsc{Loc}                     & 50  & 92.75 & 94.25 & 94.51 & 89.40 & 92.92 & 94.10 & 92.98 & 94.48 & 94.94 & 95.75 & 89.99 & 95.26 & 93.44  \\
\textsc{Loc}                     & 100 & 92.66 & 94.23 & 94.47 & 89.37 & 92.92 & 94.04 & 93.06 & 94.45 & 94.92 & 95.70 & 90.03 & 95.22 & 93.43  \\ \midrule

\textsc{CRF2o}                   & 4   & 93.46 & 94.07 & 93.97 & 89.43 & 93.03 & 93.97 & 93.08 & 94.72 & 94.82 & 95.49 & 90.19 & 94.94 & 93.43  \\
\textsc{CRF2o}                   & 50  & 93.17 & 94.05 & 93.19 & 89.35 & 93.06 & 93.93 & 93.08 & 94.67 & 94.65 & 95.47 & 90.13 & 94.89 & 93.30  \\
\textsc{CRF2o}                   & 100 & 93.03 & 94.00 & 93.91 & 89.39 & 92.92 & 93.91 & 93.08 & 94.63 & 94.65 & 95.47 & 90.13 & 94.88 & 93.33  \\ \midrule

\textsc{ArcLoc} 0T & 4   & 92.64 & 93.98 & 94.51 & 88.66 & 92.70 & 93.78 & 92.98 & 94.33 & 94.74 & 95.60 & 89.86 & 95.19 & 93.25  \\
\textsc{ArcLoc} 0T & 50  & 93.14 & 94.28 & 94.62 & 89.18 & 92.96 & 94.11 & 93.12 & 94.59 & 95.03 & 95.83 & 90.15 & 95.34 & 93.53  \\
\textsc{ArcLoc} 0T & 100 & 93.21 & 94.34 & \textbf{94.65} & 89.34 & 93.03 & 94.20 & 93.17 & 94.61 & 94.97 & 95.79 & 90.20 & 95.36 & 93.57  \\ \midrule

\textsc{ArcLoc} 1T  & 4   & 93.19 & 94.18 & 94.51 & 88.82 & 92.87 & 93.94 & 93.11 & 94.40 & 94.88 & 95.72 & 90.03 & 95.19 & 93.40  \\
\textsc{ArcLoc} 1T  & 50  & 93.51 & 94.48 & 94.63 & 89.42 & 93.09 & 94.23 & \textbf{93.23} & 94.63 & 95.13 & 95.94 & 90.22 & 95.34 & 93.66  \\
\textsc{ArcLoc} 1T  & 100 & \textbf{93.67} & \textbf{94.51} & 94.60 & \textbf{89.49} & \textbf{93.15} & 94.32 & \textbf{93.23} & \textbf{94.79} & \textbf{95.14} & \textbf{95.99} & \textbf{90.30} & \textbf{95.38} & \textbf{93.71}  \\ \midrule

\textsc{ArcLoc} 2T & 4   & 93.06 & 94.19 & 94.49 & 88.86 & 92.88 & 93.98 & 93.05 & 94.47 & 94.84 & 95.82 & 89.99 & 95.20 & 93.40  \\
\textsc{ArcLoc} 2T & 50  & 93.53 & 94.49 & 94.62 & 89.40 & \textbf{93.15} & 94.28 & 93.19 & 94.63 & 95.06 & 95.94 & 90.26 & 95.35 & 93.66  \\
\textsc{ArcLoc} 2T & 100 & \textbf{93.67} & \textbf{94.51} & 94.63 & 89.46 & 93.14 & \textbf{94.36} & 93.21 & 94.72 & \textbf{95.14} & 95.98 & 90.27 & 95.36 & 93.70  \\ \midrule
\end{tabular}
\end{adjustbox}
\caption{Dev LAS for 12 languages in UD2.2 for different numbers of parameters per model and different numbers of layers for \textsc{ArcLoc}
}\label{tab:ud-appendix}
\end{table*}

We report UD dev set results using gold POS in Table~\ref{tab:ud-appendix}.
In this case, we see that \textsc{ArcLoc} struggles to improve over \textsc{Loc} in the 4M regime, and that adding more allows parameters \textsc{ArcLoc} to recover the performance gap, while it has a detrimental effect on \textsc{Loc}.
Adding transformer layers for arc representation refinement is useful in this setting, especially in big and large settings.

\section{Error Analysis: French and English UD Treebanks}\label{sec:error_analysis}

This section provides a comparative analysis of the error rates across the French and English Universal Dependencies (UD) treebanks for the three parsing systems: \textsc{Loc}, \textsc{ArcLoc 0T}, and \textsc{ArcLoc 1T}. We analyze errors based on attachment distance, depth in the tree, part-of-speech (POS) tags, specific words, and dependency relations. The error trends and insights are discussed for both languages.

\subsection{Error Rates for Words with Different Error Rates Across Systems}

In this subsection, we analyze the words where one parsing system has error rates that are at least three times higher than another system. This comparison highlights significant performance differences between the systems when parsing certain words, emphasizing areas where certain models underperform.

\begin{figure}
    \centering
    \includegraphics[width=1\columnwidth]{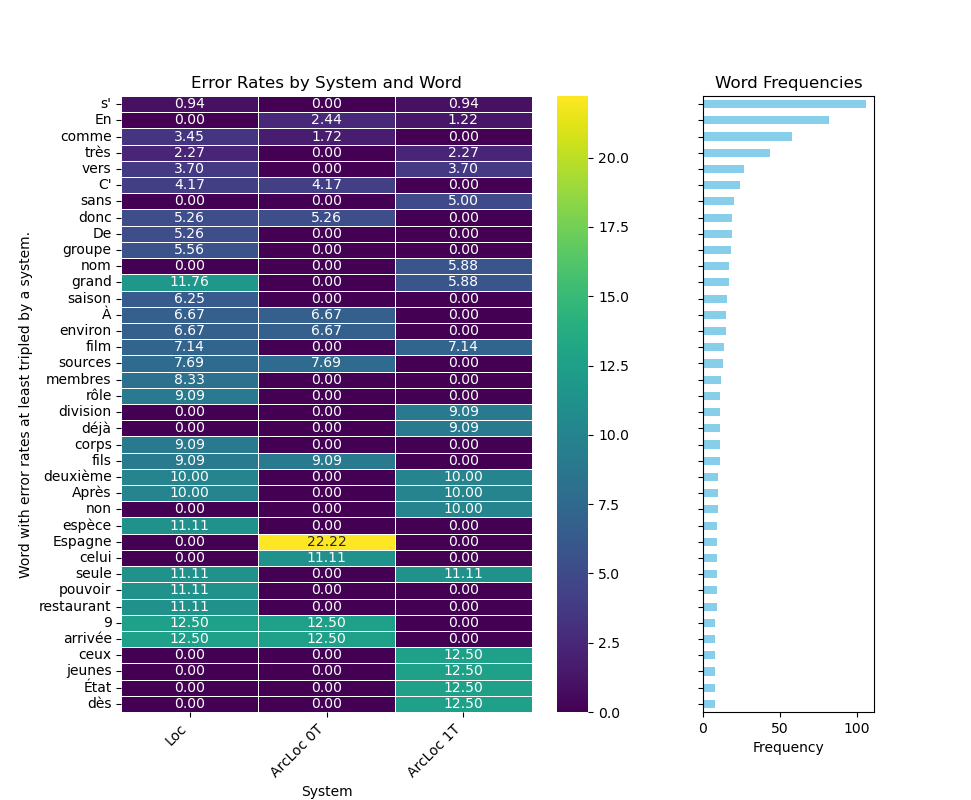}
    \caption{French error rates for words where one system has at least three times the error rate of another.}
    \label{fig:french_word_error_trippled}
\end{figure}

Figure~\ref{fig:french_word_error_trippled} shows the error rates for French words where one system has at least three times the error rate of another system. In the French dataset, words such as \emph{Espagne} and \emph{grand} exhibit large disparities between systems. For example, \textsc{ArcLoc 0T} struggles significantly more with the word \emph{Espagne}, recording an error rate of 22.22\%, whereas both \textsc{Loc} and \textsc{ArcLoc 1T} make no errors. Similarly, the word \emph{grand} shows high error rates for \textsc{Loc}, with an error rate of 11.76\%, while \textsc{ArcLoc 0T} and \textsc{ArcLoc 1T} have much lower error rates.

\begin{figure}
    \centering
    \includegraphics[width=1\columnwidth]{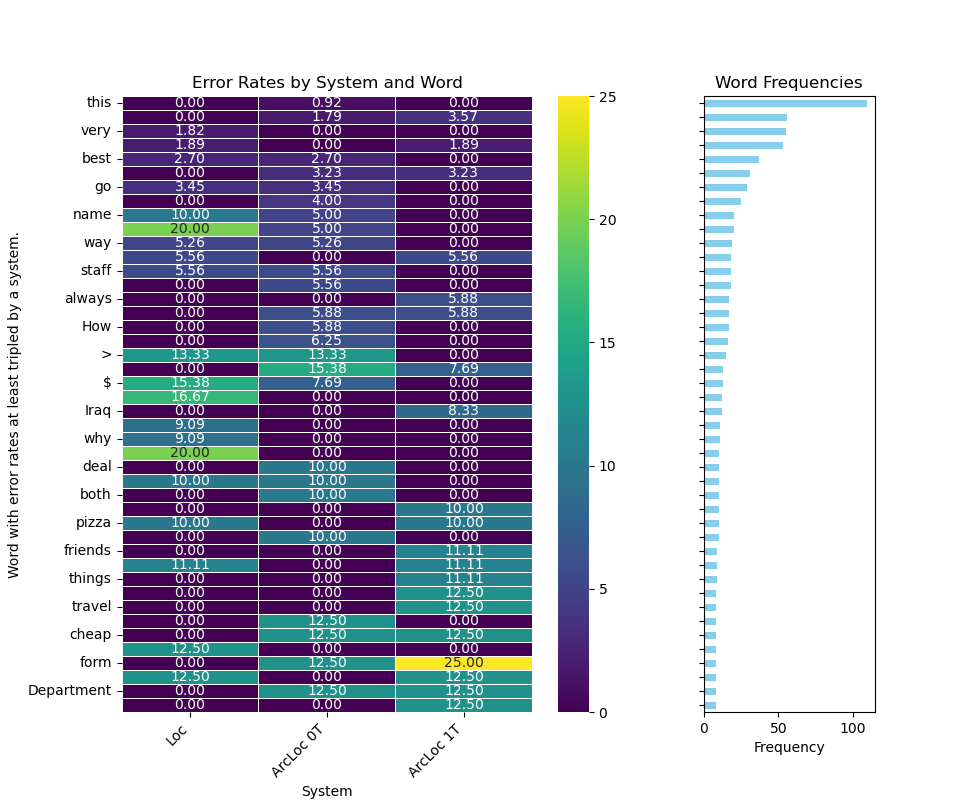}
    \caption{English error rates for words where one system has at least three times the error rate of another.}
    \label{fig:english_word_error_trippled}
\end{figure}

Figure~\ref{fig:english_word_error_trippled} provides a similar comparison for the English dataset. Words like \emph{form} and \emph{Department} show stark differences in performance.

These discrepancies are likely due to challenges in handling certain lexical or syntactic constructions.

\begin{figure}
    \centering
    \includegraphics[width=1\columnwidth]{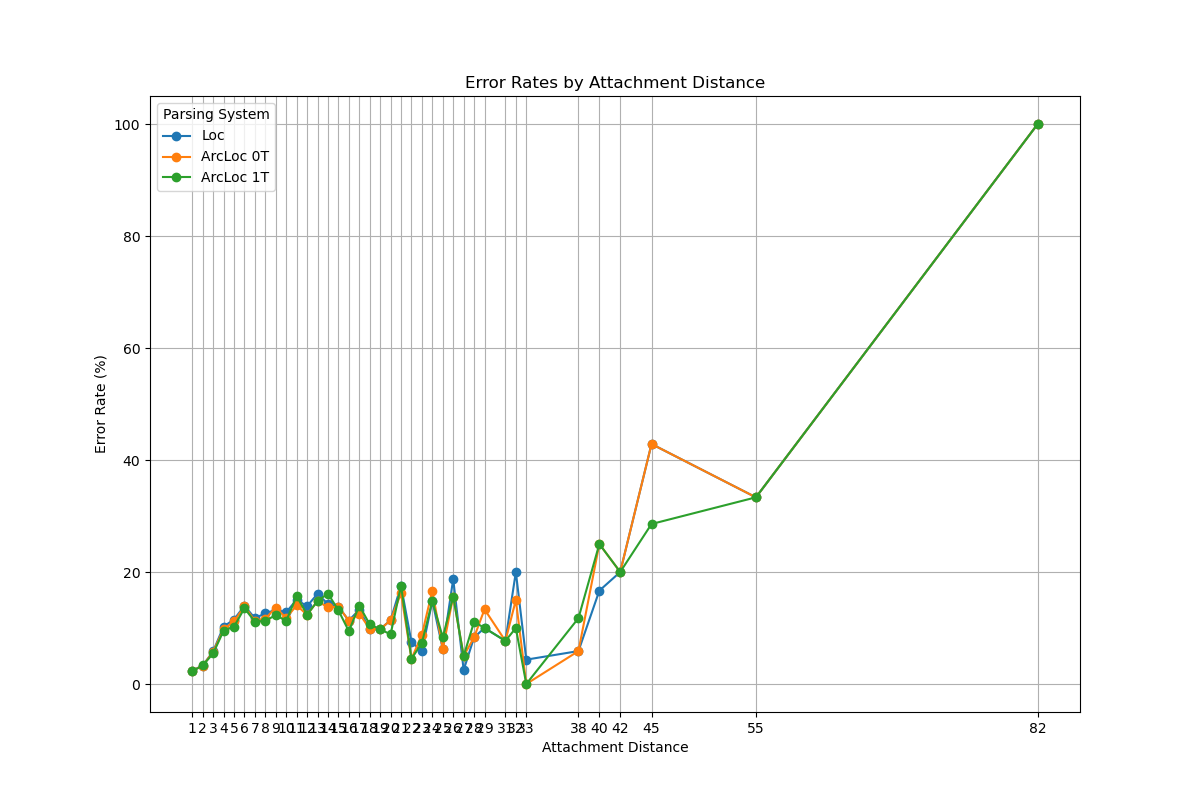}
    \caption{French error rates by attachment distance.}
    \label{fig:french_attachment_distance}
\end{figure}

\subsection{Error Rates by Attachment Distance}
\begin{figure}
    \centering
    \includegraphics[width=1\columnwidth]{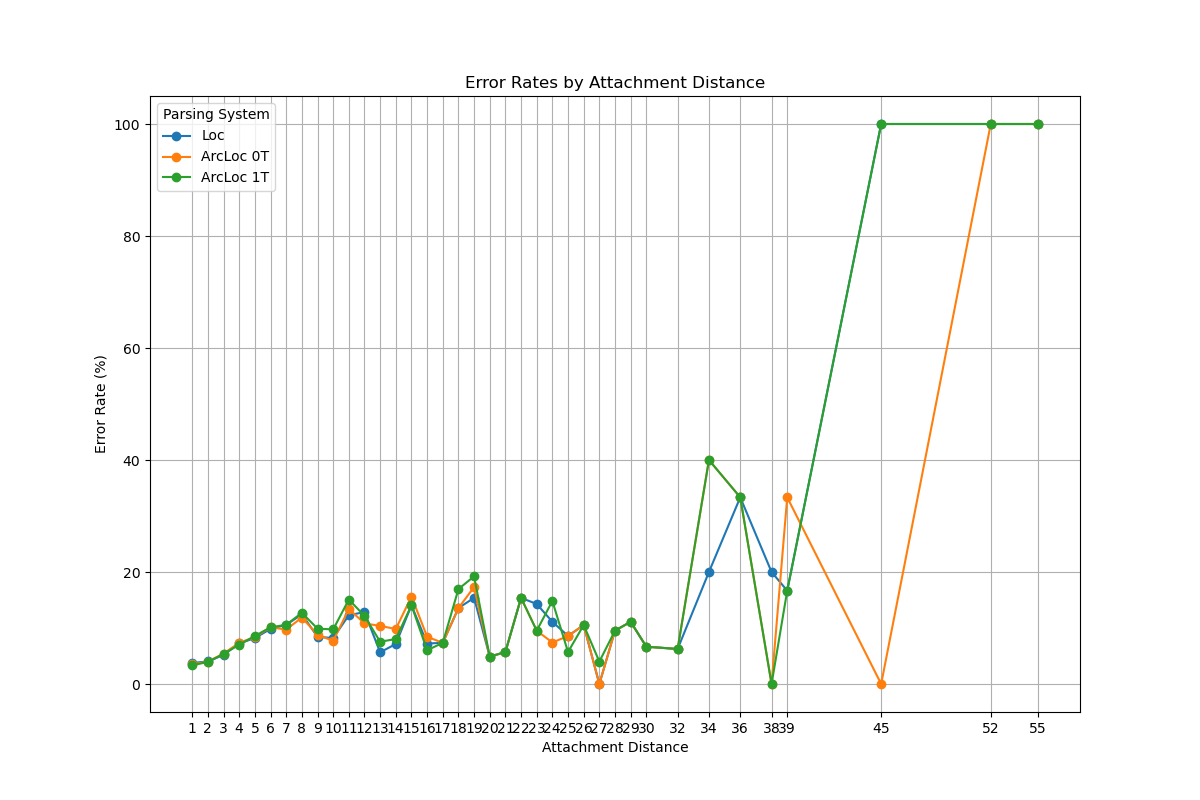}
    \caption{English error rates by attachment distance.}
    \label{fig:english_attachment_distance}
\end{figure}

Figures~\ref{fig:french_attachment_distance} and~\ref{fig:english_attachment_distance} show the error rates as a function of attachment distance for French and English, respectively. For both languages, the systems perform well on short attachment distances (below 20), with error rates staying below 20\%. However, as the attachment distance increases, the performance diverges. In French, \textsc{ArcLoc 1T} shows a steep increase in error rates beyond distance 30, while in English, \textsc{ArcLoc 0T} exhibits a sharp rise at distances above 40. These findings suggest that handling long-distance dependencies remains a challenge for all systems, particularly in French, where the errors rise more rapidly at shorter distances.

\subsection{Error Rates by POS Tags}
\begin{figure}
    \centering
    \includegraphics[width=1\columnwidth]{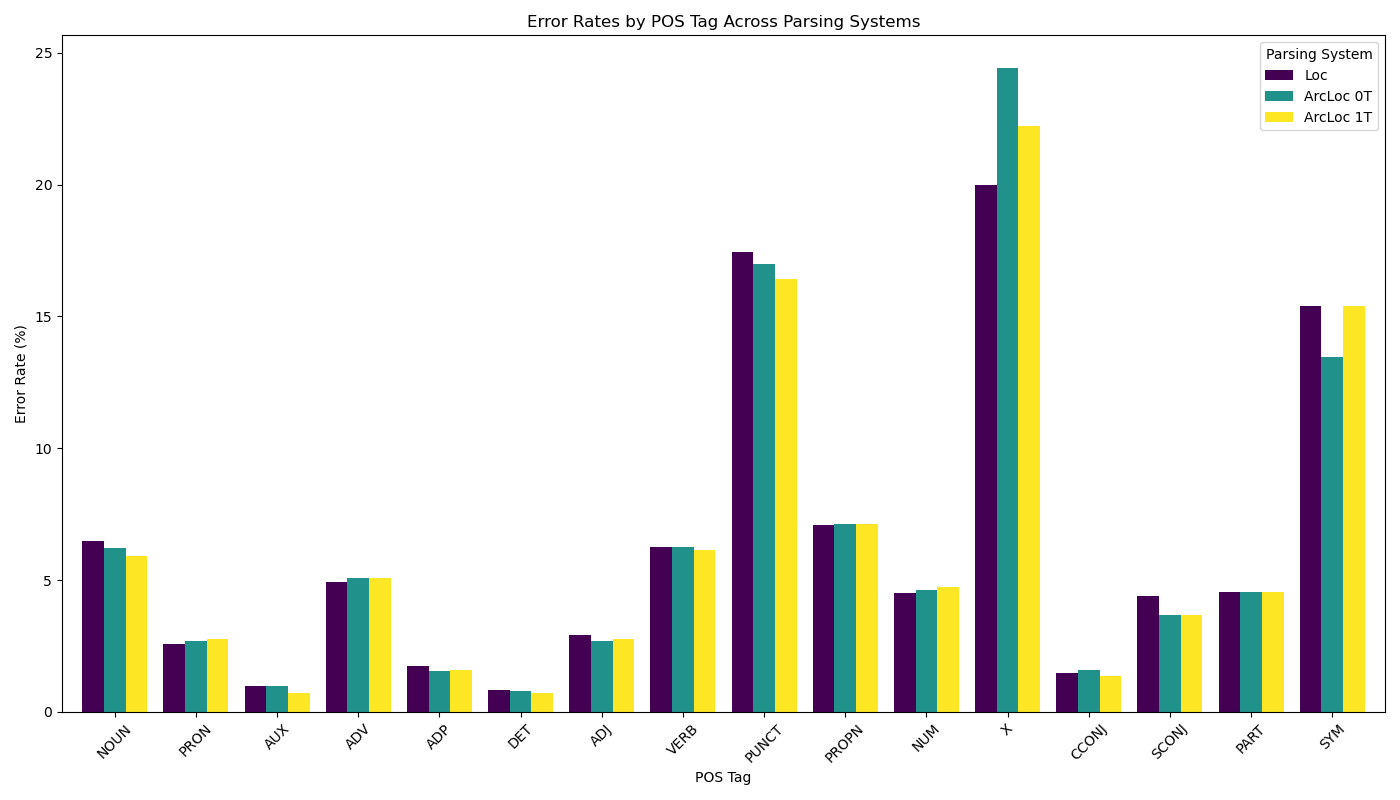}
    \caption{French error rates by POS tags.}
    \label{fig:french_pos_tags}
\end{figure}

\begin{figure}
    \centering
    \includegraphics[width=1\columnwidth]{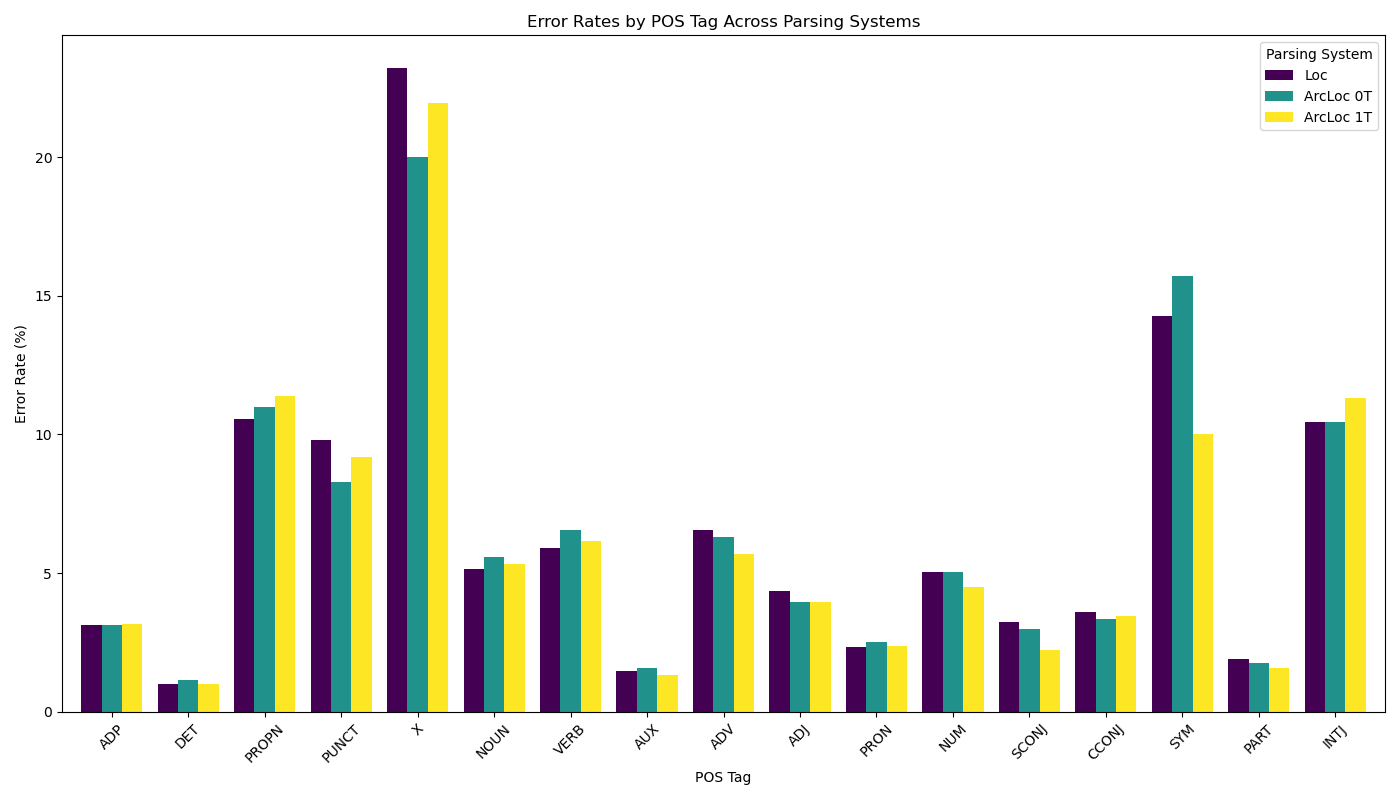}
    \caption{English error rates by POS tags.}
    \label{fig:english_pos_tags}
\end{figure}

Figures~\ref{fig:french_pos_tags} and~\ref{fig:english_pos_tags} display the error rates across different POS tags for French and English. Both languages exhibit similar trends, with the highest error rates found for punctuation (PUNCT) and unknown symbols (X). For content words like nouns (NOUN) and verbs (VERB), the systems show relatively low error rates (below 10\%). However, function words like pronouns (PRON), symbols (SYM), and conjunctions (CCONJ) are prone to higher error rates. The systems show higher sensitivity to these categories in English, particularly for SYM and INTJ, where errors exceed 20\%.

\subsection{Error Rates by Depth in the Tree}
\begin{figure}
    \centering
    \includegraphics[width=1\columnwidth]{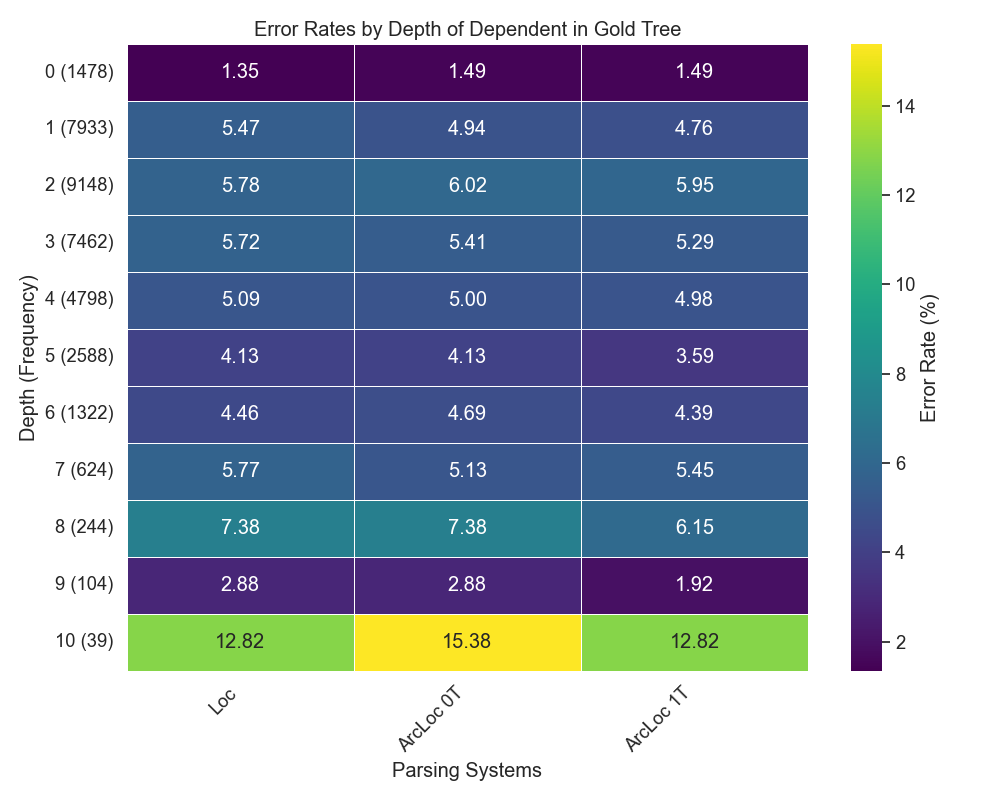}
    \caption{French error rates by depth of dependent in the tree.}
    \label{fig:french_depth_in_tree}
\end{figure}

\begin{figure}
    \centering
    \includegraphics[width=1\columnwidth]{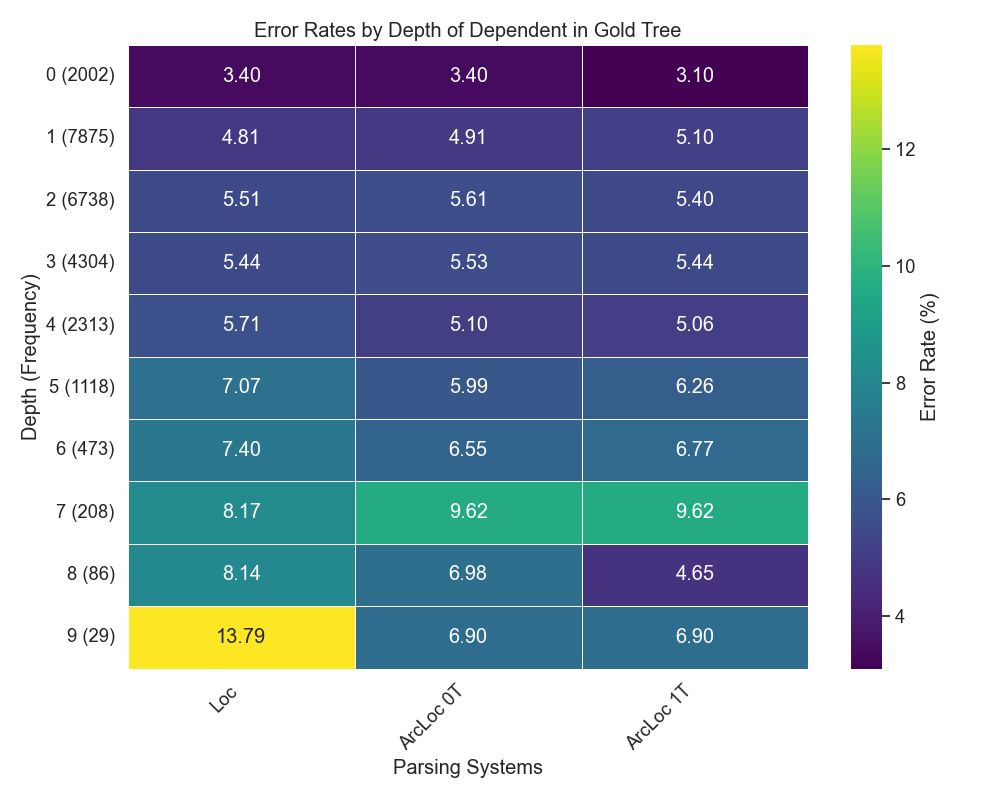}
    \caption{English error rates by depth of dependent in the tree.}
    \label{fig:english_depth_in_tree}
\end{figure}

Figures~\ref{fig:french_depth_in_tree} and~\ref{fig:english_depth_in_tree} present the error rates by depth of the dependent in the tree. For both languages, error rates are relatively low for shallow dependencies (depths 0 to 4). However, as depth increases, so do the error rates. In both French and English, \textsc{Loc} performs slightly worse at deeper levels, with error rates reaching up to 13.79\% for depth 9 in French, and around 16\% for depth 7 in English. In general, the deeper the dependency, the harder it is for all systems to maintain accuracy, with \textsc{ArcLoc 0T} performing somewhat better at deeper levels in English compared to French.

\subsection{Error Rates by Dependency Relations}
\begin{figure}
    \centering
    \includegraphics[width=1\columnwidth]{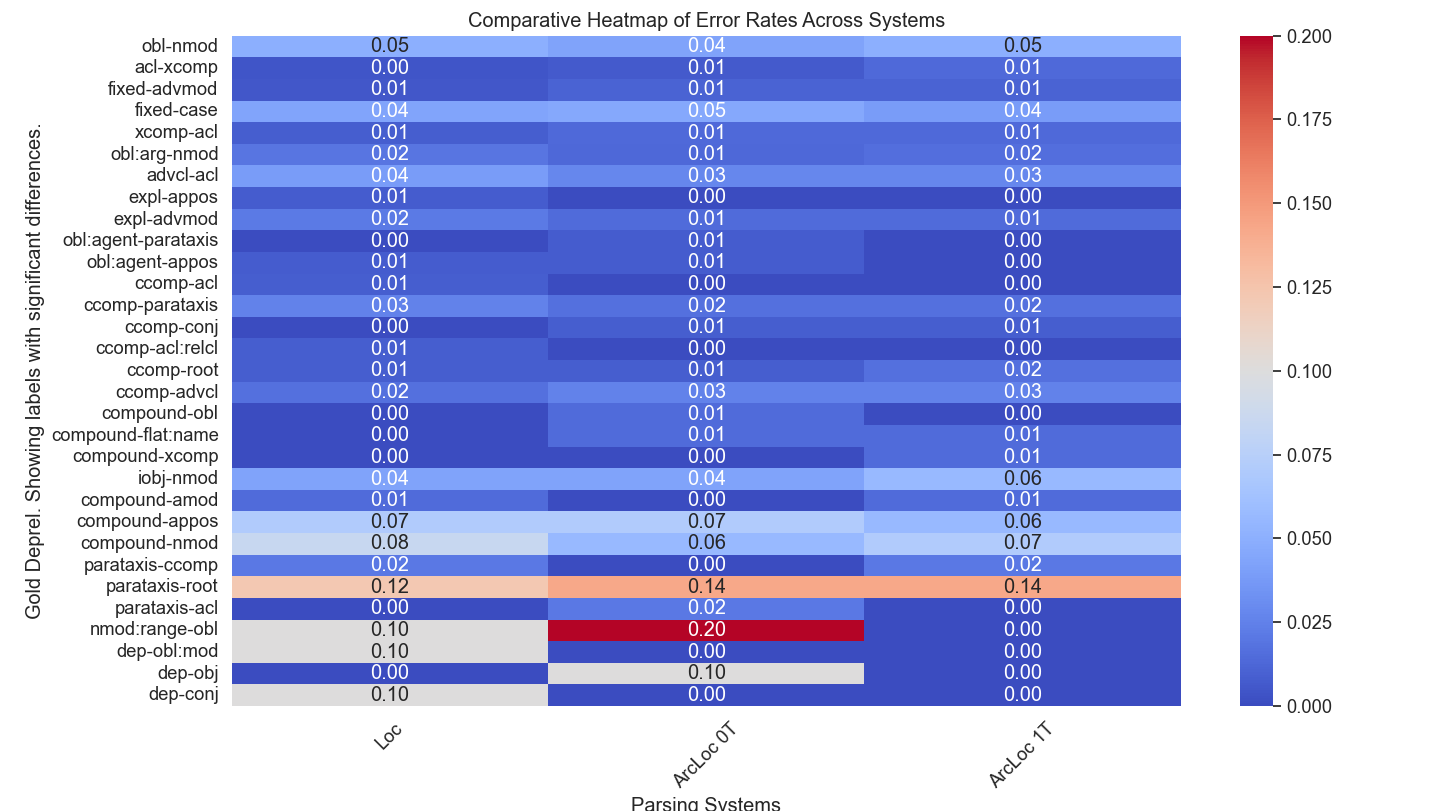}
    \caption{French heatmap of error rates by dependency relations.}
    \label{fig:french_deprel_errors}
\end{figure}

\begin{figure}
    \centering
    \includegraphics[width=1\columnwidth]{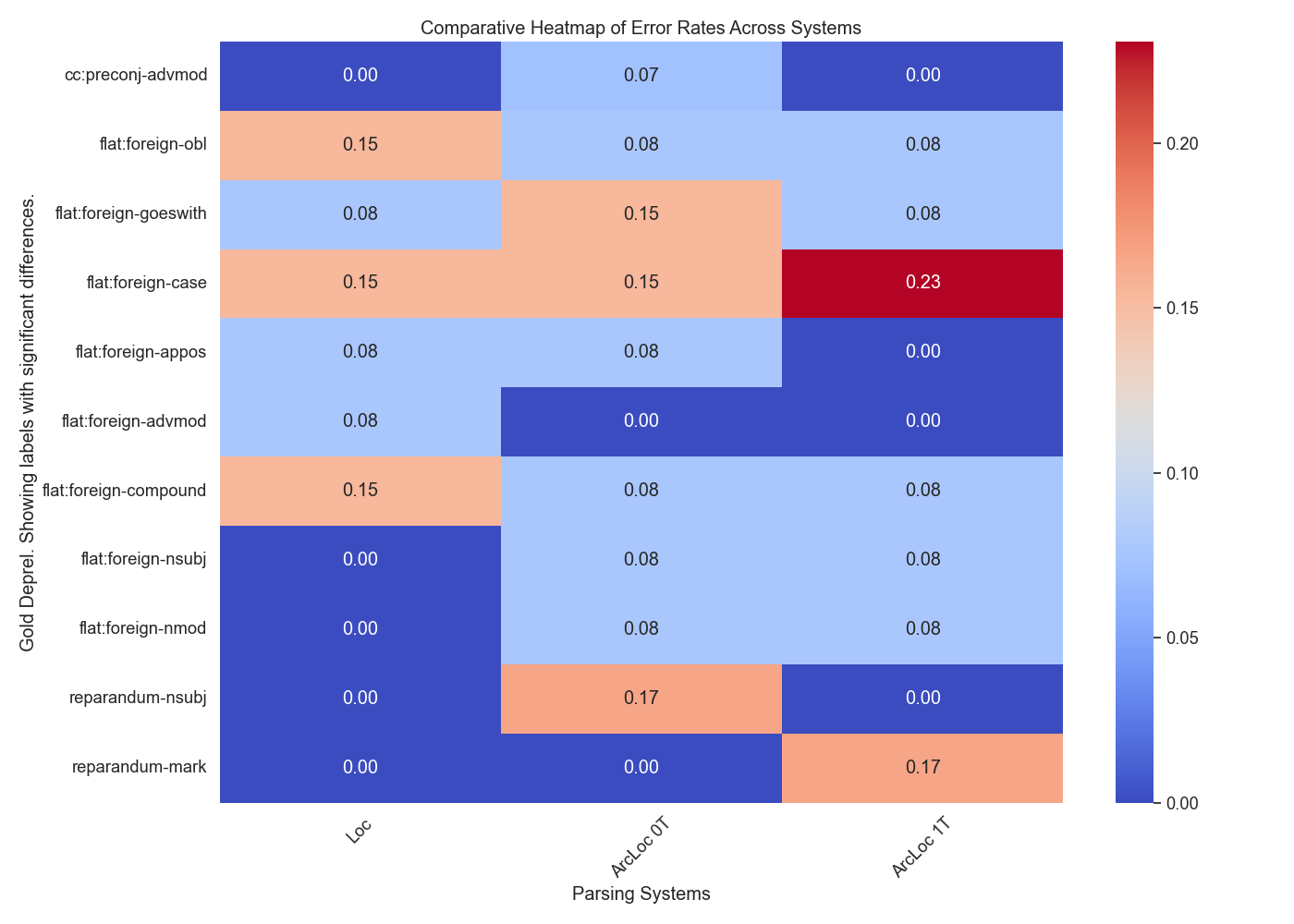}
    \caption{English heatmap of error rates by dependency relations.}
    \label{fig:english_deprel_errors}
\end{figure}

Figures~\ref{fig:french_deprel_errors} and~\ref{fig:english_deprel_errors} present heatmaps of error rates across different dependency relations for French and English. In both languages, complex relations like \texttt{parataxis-root} and \texttt{nmod:obl} exhibit the highest error rates. While \textsc{ArcLoc 0T} shows higher errors for French in these challenging relations, it performs better on average for English, especially in long-distance relations such as \texttt{flat:foreign-compound} and \texttt{fixed-case}. This indicates that while certain syntactic structures are universally challenging, language-specific factors also contribute to system performance differences.

\subsection{Raw Error Counts by Distance from Head}
\begin{figure}
    \centering
    \includegraphics[width=1\columnwidth]{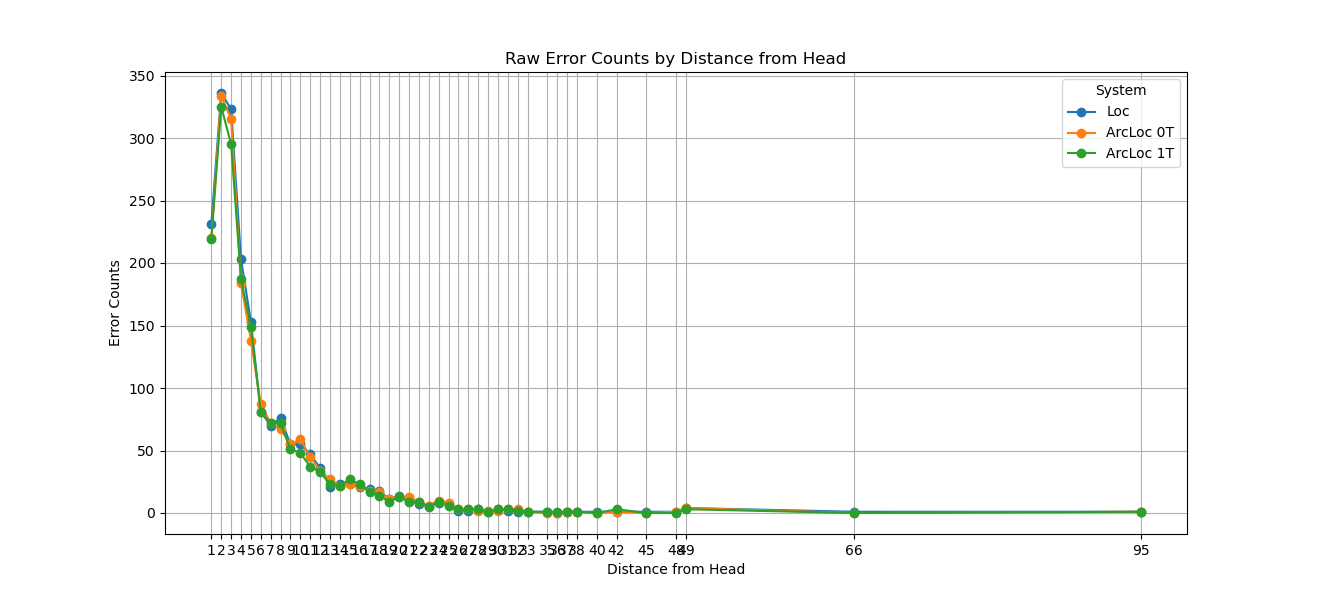}
    \caption{French raw error counts by distance from head.}
    \label{fig:french_raw_errors_distance}
\end{figure}

\begin{figure}
    \centering
    \includegraphics[width=1\columnwidth]{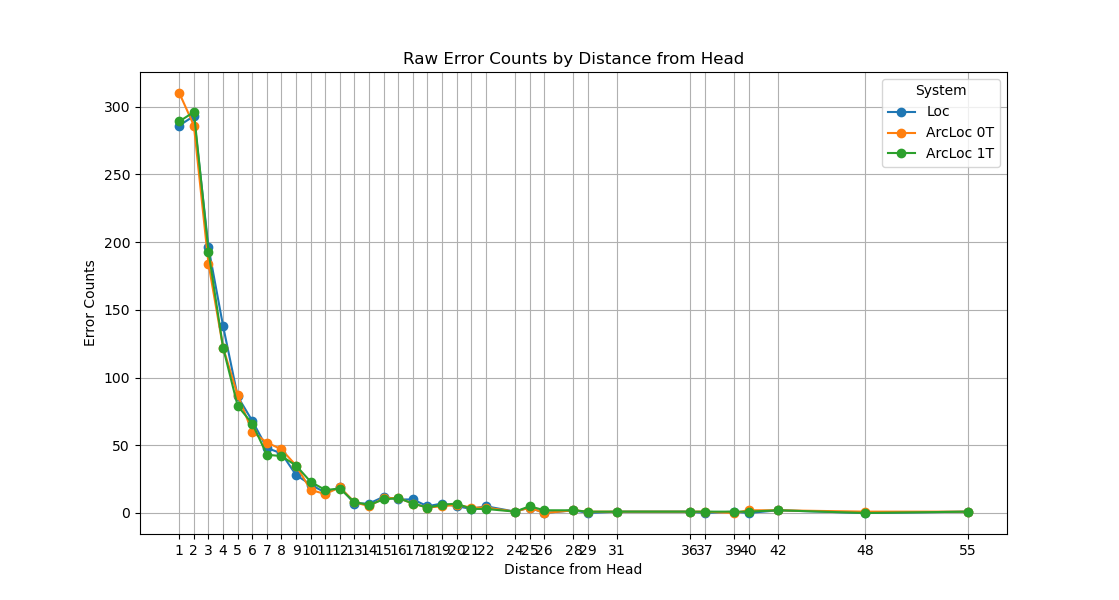}
    \caption{English raw error counts by distance from head.}
    \label{fig:english_raw_errors_distance}
\end{figure}

Figures~\ref{fig:french_raw_errors_distance} and~\ref{fig:english_raw_errors_distance} present the raw error counts as a function of distance from the head. For both languages, the majority of errors occur at short distances (1 to 5 words), where dependency relations are the most frequent. The error count decreases as the distance increases, but significant spikes in errors occur beyond distance 30, particularly in French. This confirms that handling long-range dependencies remains a common challenge across both languages and all parsing systems.

\end{document}